\documentclass[runningheads]{llncs}

\usepackage[utf8]{inputenc}
\usepackage{graphicx}
\usepackage{amsmath,amssymb}
\usepackage{array}
\usepackage{color,soul}
\usepackage{algorithm}
\usepackage{algpseudocode}
\usepackage{siunitx}
\usepackage{booktabs}
\usepackage{svg}

\usepackage[font={footnotesize}, labelfont=bf, labelsep=period]{caption}
\newif\ifdraft
\drafttrue

\definecolor{orange}{rgb}{1,0.5,0}
\definecolor{pink}{rgb}{0.98, 0.38, 0.5}

\ifdraft
 \newcommand{\RS}[1]{{\color{red}{\bf RS: #1}}}
 
 \newcommand{\PMN}[1]{{\color{orange}{\bf PMN: #1}}}

 \newcommand{\MH}[1]{{\color{pink}{\bf MH: #1}}}

\else
 \newcommand{\RS}[1]{{\color{red}{}}}
 
 \newcommand{\PMN}[1]{{\color{red}{}}}
 
  \newcommand{\MH}[1]{{\color{red}{}}}
 
\fi

\newcommand{\real}{\mathbb{R}}
\newcommand{\X}{\mathbf{X}}
\newcommand{\Y}{\mathbf{Y}}
\newcommand{\D}{\mathbf{D}}

\newcommand{\h}{\mathbf{h}}

\newcommand{\comment}[1]{}

\newcommand{\eg}{e.\,g.,\ }
\newcommand{\ie}{i.\,e.,\ }

\begin{document}

\title{Fused Detection of Retinal Biomarkers in \\OCT Volumes}
\author{Thomas Kurmann\inst{1} \and Pablo Márquez-Neila\inst{1} \and Siqing Yu\inst{2} \and Marion Munk\inst{2} \and Sebastian Wolf\inst{2} \and Raphael Sznitman\inst{1}}
\authorrunning{T. Kurmann et al.}
% index{Kurmann, Thomas}
% index{Márquez-Neila, Pablo}
% index{Yu, Siqing}
% index{Munk, Marion}
% index{Wolf, Sebastian}
% index{Sznitman, Raphael}
\institute{
University of Bern, Bern, Switzerland
\and
University Hospital of Bern, Bern,  Switzerland
}

\maketitle

\begin{abstract}
Optical Coherence Tomography (OCT) is the primary imaging modality for detecting pathological biomarkers associated to retinal diseases such as Age-Related Macular Degeneration. In practice, clinical diagnosis and treatment strategies are closely linked to biomarkers visible in OCT volumes and the ability to identify these plays an important role in the development of ophthalmic pharmaceutical products. In this context, we present a method that automatically predicts the presence of biomarkers in OCT cross-sections by incorporating information from the entire volume. We do so by adding a bidirectional LSTM to fuse the outputs of a Convolutional Neural Network that predicts individual biomarkers. We thus avoid the need to use pixel-wise annotations to train our method and instead provide fine-grained biomarker information regardless. On a dataset of 416 volumes, we show that our approach imposes coherence between biomarker predictions across volume slices and our predictions are superior to several existing approaches.

\end{abstract}

% !TEX root = paper1547_top.tex
% !TEX spellcheck = en-US

\section{Introduction}
\label{sec:intro}
\begin{figure}
\centering
\includegraphics[width=0.8\textwidth]{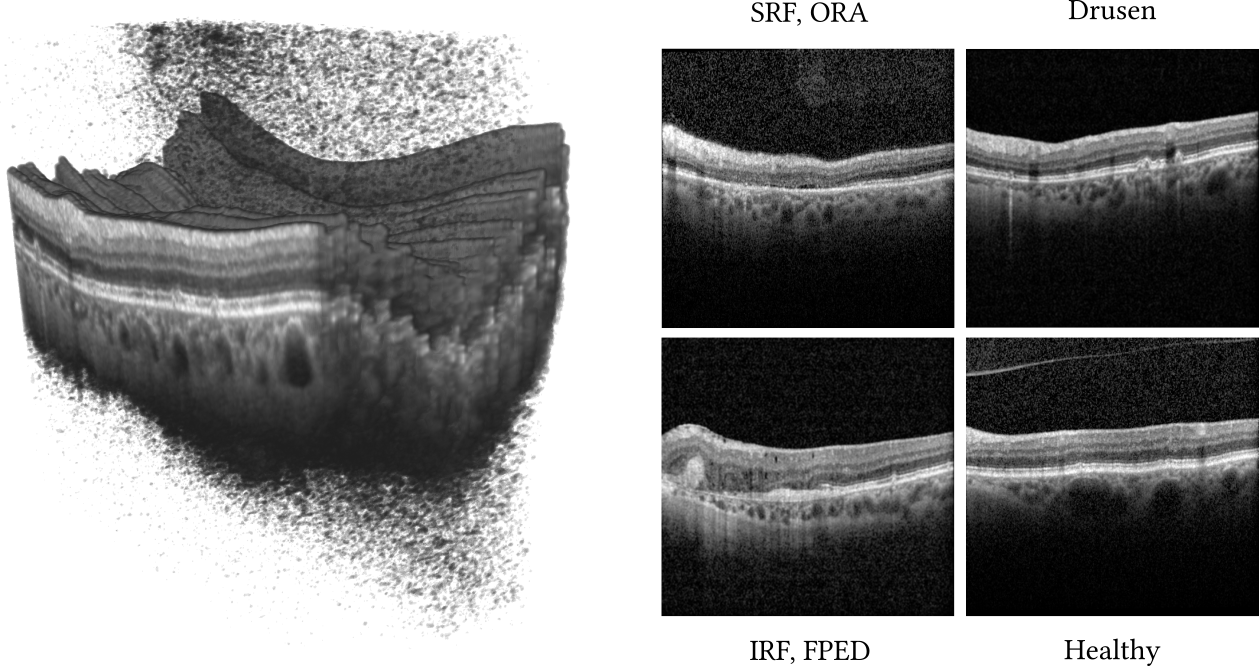}
\caption{(\textbf{left}) Example of an OCT volume of a patient with AMD. Slices are misaligned, even after post-processing registration. (right) four slices extracted from the volume, each containing a different set of biomarkers indicated for each image. }
\label{fig:examples}
\end{figure}

Age-Related Macular Degeneration (AMD) and Diabetic Macular Edema (DME) are
chronic sight-threatening conditions that affect over 250 million people
world wide~\cite{Bourne2017}. To diagnose and manage these diseases, Optical
Coherence Tomography (OCT) is the standard of care to image the retina safely and quickly (see
Fig.~\ref{fig:examples}). However, with a growing global patient population and over 30 million
volumetric OCT scans acquired each year, the resources needed to assess these has
already surpassed the capacity of knowledgeable experts to do so~\cite{Bourne2017}.

For ophthalmologists, identifying biological markers of the retina, or {\it biomarkers}, plays a critical role in both clinical routine and research. Biomarkers can include the
presence of different types of fluid buildups in the retina, retinal shape and
thickness characteristics, the presence of cysts, atrophy or scar tissue. Beyond this,
biomarkers are paramount to assess disease severity in clinical routine and
have a major role in the development of new pharmaceutical therapeutics. With over a dozen clinical and research biomarkers,
their identification is both challenging and time consuming due to their number, size, shape and extent.

To support clinicians with OCT-based diagnosis, numerous automated methods have
attempted to segment and identify specific biomarkers from OCT scans. For instance, retinal
layer~\cite{Apostolopoulos2017,Hussain2017,Roy2017} and fluid~\cite{Roy2017}
segmentation, as well as drusen detection~\cite{Zhao2017} have previously been
proposed. While these methods perform well, they are limited in the number of
biomarkers they consider at a time and often use pixel-wise annotations to train
supervised machine learning frameworks. Given the enormous annotation task involved
in manually segmenting volumes, they are often trained and evaluated on relatively small amounts
of data (20 to 100 volumes)~\cite{Apostolopoulos2017,Roy2017,Bogunovic2019-nu}.

\comment{
To avoid this, several recent works have instead focused on predicting
pathology or disease severity directly from volumes with no training
information other than the associated pathology annotation~
\cite{Hussain2017,Apostolopoulos2016,Schlegl2017,Venhuizen2017}. Similar to fine-
grain image tagging, whereby each slice within a volume must be tagged if a given
biomarker is present, these works focus on unique biomarker detection as apposed to segmentations. However, estimating much larger numbers of biomarkers (\eg $>10$) throughout a volume is far more challenging as directly infering these using traditional 3D approaches (\eg 3D Convolutional Neural Network (CNN)~\cite{xxx}) implies learning prediction functions with large output spaces (\ie $2^{50}$).
}

Instead, we present a novel strategy that automatically identifies the presence of a wide range of
biomarkers throughout an OCT volume. Our method does not require biomarker segmentation
annotations, but rather biomarker tags as to which are present on a given OCT slice. Using a large dataset of OCT slices and annotated tags, our approach then estimates all
biomarkers on each slice of a new volume separately. We do this first seperately, without considering
adjacent slices, as these are typically highly anisotropic and not aligned within the volume. We
then treat these predictions as sufficient statistics for each slice and impose biomarker coherence
across slices using a bidirectional Long short-term memory (LSTM) neural network~\cite{10.1007/11550907_126}. By doing
so, we force our network to learn the wanted biomarker co-dependencies within the volume from
slice predictions only, so as to avoid dealing with anisotropic and non-registered slices common in OCT volumes. We show in
our experiments that this leads to superior performances over a number of existing methods.

% !TEX root = paper1547_top.tex
% !TEX spellcheck = en-US

\section{Method}
\label{sec:method}
We describe here our approach for predicting biomarkers across
all slices in an OCT volume. Formally, we wish to predict the presence of $B$
different biomarkers in a volume using a deep network, $f:[0, 1]^{S\times{}W
\times{}H} \to [0, 1]^{S\times{}B}$, that maps from a volume of $S$ slices, $\X\in[0, 1]^{S
\times{}W\times{}H}$, to a set of predicted probabilities $\hat{\Y}\in[0, 1]^{S\times{}B}$. We denote $\hat\Y_{sb}$ as the estimated probability that biomarker $b$ occurs in slice~$s$.

While there are many possible network architectures for $f$, one
simple approach would be to express $f$ as $S$ copies
of the same CNN, $f':[0, 1]^{W\times{}H} \to [0, 1]^{B}$, whereby each slice in the volume is
individually predicted. However, such an architecture ignores the fact that
biomarkers are deeply correlated across an entire volume. The
other extreme would be to define $f$ as a single 3D CNN. Doing so however would be difficult
because (1) 3D~CNNs assume spatial coherence in their convolutional layers and (2) the output
of $f$ would be of dimension $[0,1]^{S \times B}$. While (1) strongly
violates OCT volume structure because there are they typically display non-rigid transformations between consecutive OCT slices, (2) would imply training with an enormous amount of training data.
% As we show in our experiments, just using the simple CNN or MLP fusion does not perform optimally in practice.

\begin{figure}[t]
\centering
\includegraphics[width=0.99\textwidth]{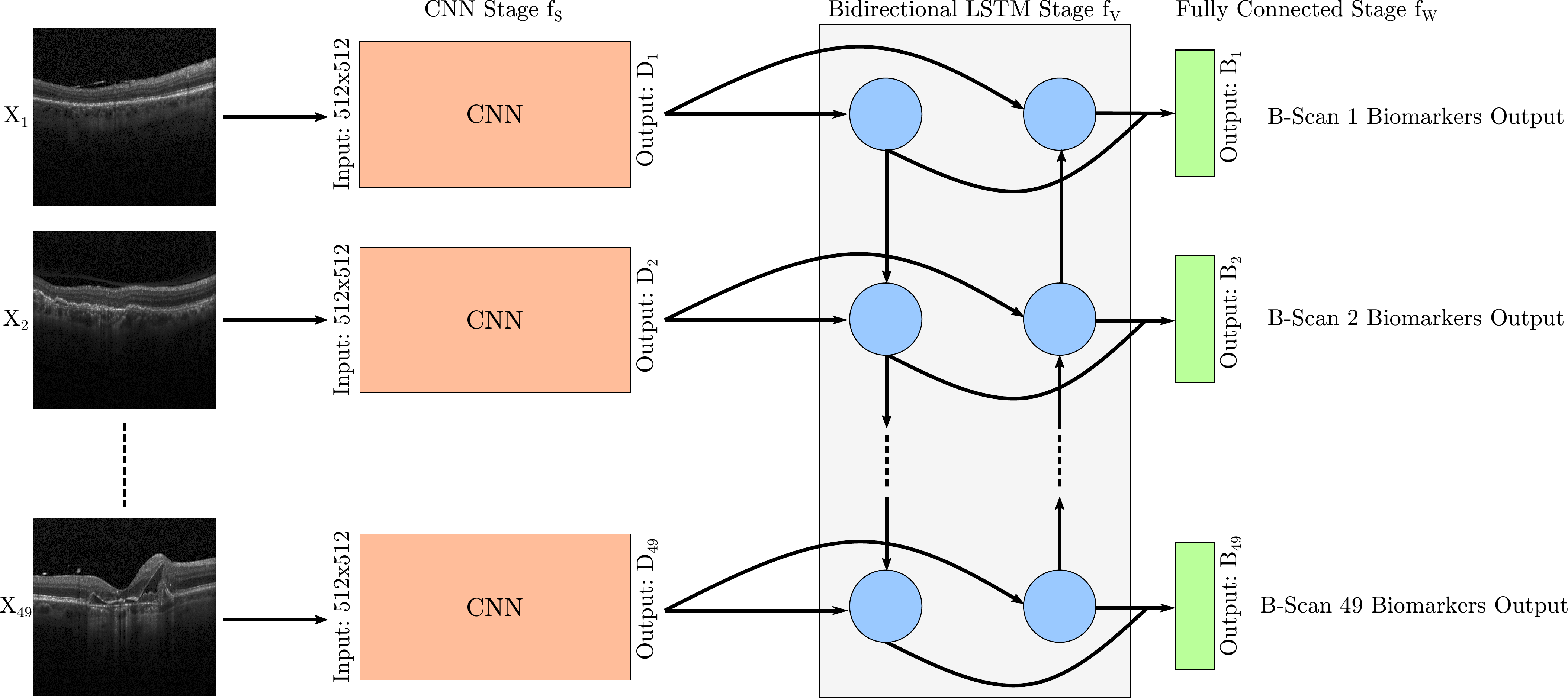}
\caption{Overview diagram of our apporach including the CNN, $f_\mathcal{S}$, the bidirectional LSTM, $f_\mathcal{V}$, and the output of the fully connected stage. OCT volumes consist of 49 slices. }
\label{fig:overview}
\end{figure}

For these reasons, we take an intermediate approach between the above mentioned
extremes and express our network as a composition $f=f_\mathcal{V}\circ{}f_\mathcal{S}$,
where $f_\mathcal{S}:[0, 1]^{S\times{}W\times{}H} \to \real^{S\times{}D}$ processes
slices individually and produces a $D$-dimensional descriptor for
each slice. Then, $f_\mathcal{V}:\real^{S\times{}D} \to [0, 1]^{S\times{}B}$ fuses all
$S$~slice descriptors and predicts the biomarker probabilities for each slice, whereby taking into account the information of the entire volume. Fig.~\ref{fig:overview} depicts our framework and we detail each of its components in the subsequent sections.

\subsection{Slice network~$f_\mathcal{S}$}
When presented with a volume~$\X$, $f_\mathcal{S}$~processes each slice independently
using the same \emph{slice convolutional network},~$f'_\mathcal{S} : [0, 1]^{W\times{}H} \to \real^{D}$
that maps from a single slice to a $D$-dimensional descriptor. The output of~$f_\mathcal{S}$ is then the concatenation of the individual descriptors,
\begin{equation}
    f_\mathcal{S}(\X) = \left[f'_\mathcal{S}(\X_1), \ldots, f'_\mathcal{S}(\X_S)\right].
\end{equation}
In our experiments, we implemented~$f'_\mathcal{S}$ as the convolutional part of a Dilated Residual Network~\cite{Yu2017-gu} up to the global pooling layer.

\subsection{Volume fusion network~$f_\mathcal{V}$}
Let~$\D=f_\mathcal{S}(\X)\in\real^{S\times{}D}$ be the set of descriptors of a
volume~$\X$ computed by~$f_\mathcal{S}$. The fusion network~$f_
\mathcal{V}$ takes~$\D$ and produces
a final probability prediction~$\hat\Y$.

The most straightforward architecture
for~$f_\mathcal{V}$ would be a multilayer perception~(MLP), which is typical
after convolutional layers. However, MLPs make no assumptions about
the underlying nature of the data. Consequently, MLPs are hard to train,
requiring either huge amounts of training data or resort to aggressive data
augmentation techniques, particularly when the dimensionality of the input
space is large as in this case. More importantly, a MLP would ignore two
important aspects about~$\D$: (1) the rows of~$\D$ belong to the same
feature space that share a common distribution; (2) volumes have spatial
structure with respect to the biomarkers within them and slices that are nearby to one another have similar descriptors.

To account for this, we use an LSTM to process slices in a sequential way and implicitly leverage spatial dependencies, while performing the same
operations on every input (\ie implicitly assuming a common distribution in the
input space).
Formally, our LSTM is a network $f_{\ell} : \real^D\times{}\real^H \to \real^H$ that receives a descriptor and the previous $H$-dimensional LSTM state, to produce a new state. We use the LSTM to iteratively process the descriptors~$\D$ generating a sequence of LSTM states,
\begin{equation}
    \label{eq:lstm_forward}
    \h_{i} = f_{\ell}(\D_s, \h_{s-1}),\quad s=1,\ldots,S,
\end{equation}
where $\D_s$ is the descriptor on slice $s$. Additionally, since the underlying distribution of OCT volumes is
symmetric\footnote{Flipping the slice order in a
volume produces another statistically correct volume.},
we use the same LSTM to process the descriptors backwards,
\begin{equation}
    \label{eq:lstm_backward}
    \h'_{i} = f_{\ell}(\D_s, \h'_{s+1}),\quad s=S,\ldots,1,
\end{equation}
generating a second sequence of LSTM states. Initial states are~$\h_0$
and~$\h'_{S+1}$, respectively.

Note that at each position~$s$, $\h_s$ and~$\h'_s$ combine the information from the
current descriptor~$\D_s$ with additional information coming from neighboring slices. We then concatenate both states in a single vector and feed it to a final
fully connected layer~$f_{\omega} : \real^{2H} \to \real^B$ that computes the
estimated probabilities. The complete volume fusion network~$f_\mathcal{V}$ is the concatenation of
the outputs of~$f_{\omega}$ for all the slices:
\begin{equation}
    f_\mathcal{V}(\D) =  \left [f_{\omega}([\h_1, \h'_1
   ]),\ldots,f_{\omega}([\h_S, \h'_S])
   \right].
\end{equation}

\subsection{Training}
Training $f$ requires a dataset of $M$~annotated volumes,~$\mathcal{D}=\{(\X^{(m)},\Y^{(m)})\}_{m=1}^M$ where for each volume~$\X^{(m)}$ a set of binary labels~$\Y^{(m)}$ is provided. $\Y^{(m)}_{sb}$~is~1 if biomarker~$b$ is present in slice~$\X^{(m)}_s$. We then use the standard binary cross entropy as our loss function,
\begin{equation}
    \label{eq:loss}
    \ell(\Y,\hat\Y) = -\sum_s\sum_b (1-\Y_{sb})\log(1-\hat\Y_{sb}) + \Y_{sb}\log \hat\Y_{sb},
\end{equation}
where $\hat\Y=f(\X)$~is the estimation of our network for a given volume. The goal during training is then to minimize the expected value of the loss~$\ell$
over the training dataset~$\mathcal{D}$.

While we could perform this
minimization with a gradient-based method in an end-to-end fashion from
scratch, we found that a two-stage training procedure
% pre-training the slice network~$f'_B$ to predict
% biomarkers of individual slices
helped boost performances at test time.
In the first stage, we train the \emph{slice network}~$f'_\mathcal{S}$ alone to predict
the biomarkers of individual slices.
More specifically, we append a temporary fully connected
layer~$f_t:\real^{D}\to [0,1]^{B}$ at the end of $f'_\mathcal{S}$, and then minimize a
cross entropy loss while presenting to the network randomly sampled slices
from~$\mathcal{D}$. In the second stage, we fix the weights of~$f'_\mathcal{S}$
and minimize the loss of Eq.~\eqref{eq:loss} for the whole architecture~$f$
updating only the weights of the volume fusion network~$f_\mathcal{V}$.

% !TEX root = paper1547_top.tex
% !TEX spellcheck = en-US

\section{Experiments}
\label{sec:exp}

\subsection{Data}
Our dataset consists of 416 volumes (originating from 327 individuals with Age-Related Macula Degeneration and Diabetic Retinopathy) whereby each volume scan consists of $S=49$~slices for a total of 20'384~slices. Volumes were obtained using the Heidelberg Spectralis OCT and each OCT slice has a resolution of $496 \times 512$ pixels. Trained annotators provided slice level annotations for 11 common biomarkers: Subretinal Fluid (SRF), Interetinal Fluid (IRF), Hyperreflective Foci (HF), Drusen, Reticular Pseudodrusen (RPD), Epiretinal Membrane (ERM), Geographyic Atrophy (GA), Outer Retinal Atrophy (ORA), Intraretinal Cysts (IRC), Fibrovascular PED (FPED) and Healthy. The dataset was randomly split $90\%/10\%$ for training and testing purposes, making sure that no volume from the same individual was present in both the training and test sets. These sets contained a total of 18'179 and 2'205~slices, respectively. The distribution of biomarkers in the training and test sets are reported in Table~\ref{table:table-results}. For all our experiments, we performed 10-fold cross validation, where the training set was split into a training (90\%) and validation (10\%) set.

\subsection{Parameters and baselines}
For our approach, we set $D=512$ for the size of the $f_\mathcal{S}$ descriptors and $H=512$ for the size of the LSTM hidden state. We train the fusion stage using a batch size of 4~volumes, while training using SGD with momentum of~$0.9$ and a base learning rate of~$10^{-2}$ which we decrease after 10 epochs of no improvement in the validation score.

To demonstrate the performance of our approach, we compare it to the following baselines:
\begin{itemize}
\item {\bf Base:} the output of $f_\mathcal{S}$ (e.g. no slice fusion).
\item {\bf MLP}: output of size $49\times11$ using the $49\times512$ sized feature matrix from the {\bf Base} classifier.
\item {\bf Conv-BLSTM:}  fuses the last convolutional channels of $f_\mathcal{S}$ with a size of $(49,64,64,256)$ and a hidden state of $H=256$ channels. This is then followed by a global pooling and a fully connected layer.
\item {\bf Conditional Random Field (CRF):} trained to learn co-occurrence of biomarkers within each slice and to smooth the predictions for each biomarker along different slices of the volume. Logit outputs of the {\bf Base} classifier are used as unary terms, and learned pairwise weights are constrained to be positive to enforce submodularity of the CRF energy. We use the method from~\cite{szummer2008} for training and standard graph-cuts for inference at test time.
\end{itemize}
For all methods we use the same {\bf Base} classifier and train it as a multi-label classification task using a learning rate of~$10^{-2}$, a batch size of~32 with SGD and a momentum of~$0.9$. Rotational and flipping data augmentation was applied during training. We retain the best model for evaluation and do not perform test time augmentation or cropping. The network was pre-trained on ImageNet~\cite{ILSVRC15}.

Our primary evaluation metric are the micro and macro mean Average Precision (mAP). In addition, we also report the Exact Match Ratio (EMR) which is equal to the percentage of slices predicted without failing to detect any biomarker in it. The mAP of the CRF baseline is not directly comparable as the CRF output is binary, hence allowing only a single preciscion-recall point to be evaluated. We therefore also state the maximum F1 scores for each method.

\subsection{Results}
Table~\ref{table:table-results} reports the performances of all methods. Using the proposed method we see an increase in mAP across all biomarkers except for GA and ORA. Both biomarkers have a very low sample size in the test set. The proposed method outperforms all other fusion methods in terms of mAP and F1 score and considerably improves over the unfused baseline, which confirms our hypothesis that inter-slice dependencies can be used to increase the per slice performance. The poor performance of the Convolutional BLSTM can be explained due to the misalignment of adjacent slices.

\begin{figure}
\centering
\includegraphics[width=0.99\textwidth]{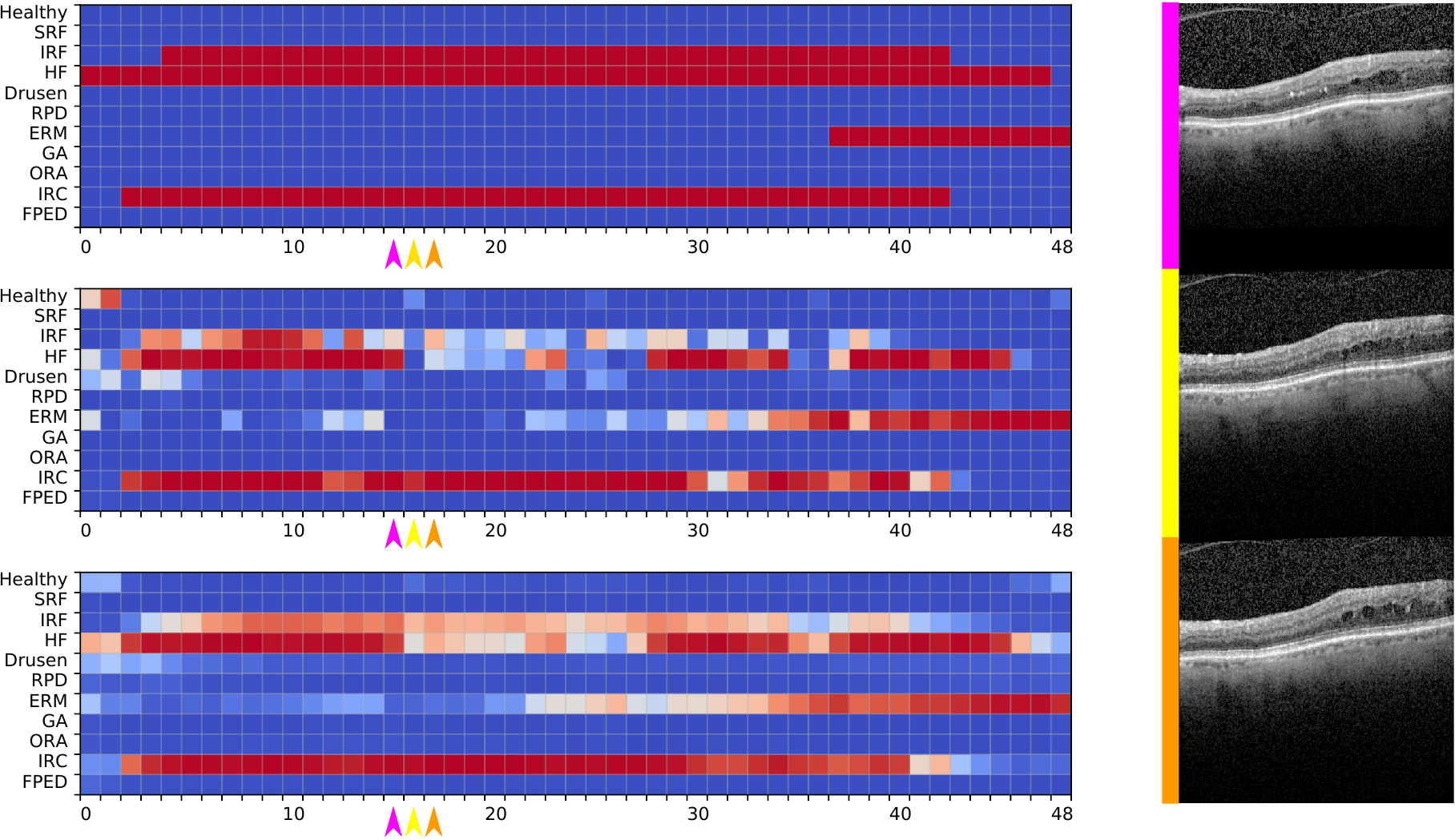}
\caption{Volume classification example, where (top left) depicts the ground truth, (middle left) the {\bf Base} classification and (bottom left) our proposed method. Warm colors indicate higher likelihood of presence. (Right) Three consecutives slices where the {\bf Base} classifier incorrectly misses biomarker IRF in the center slice (yellow). Our proposed method manages to fuse information from adjacent slices (pink and orange) to infer the proper prediction.
}
\label{fig:result_example}
\end{figure}

In Fig. \ref{fig:result_example}, we show a typical example illustrating the performance improving ability of our proposed method. In particular, we show here the prediction of our approach on each slice for each biomarker and highlight three consecutive slices of the tested volume (right). For comparison, we also show the corresponding ground-truth (top left) and the outcome from the {\bf Base} classifier (middle left). Here we see that our approach is capable of inferring more accurately the set of biomarkers across the different slices.

\begin{table}[t!]
\sisetup{
table-alignment = center,
table-column-width= 0.14\textwidth,
table-figures-integer = 5,
table-figures-decimal = 0,
}
\centering
\resizebox{\textwidth}{!}{\begin{tabular}{lcccccc}
\toprule
\textbf{Biomarker} & \textbf{Base} & \textbf{MLP} & \textbf{Conv-BLSTM} & \textbf{CRF} & \textbf{Proposed}  \\ \midrule
Healthy (5310/494) & 0.797$\pm$0.023  & 0.730$\pm$0.025  & 0.795$\pm$0.013  & - & \bf 0.800$\pm$0.022 \\
SRF (942/103) & 0.847$\pm$0.024  & 0.796$\pm$0.043  & 0.877$\pm$0.030  & - & \bf 0.905$\pm $0.017 \\
IRF (2019/339) & 0.691$\pm$0.044  & 0.705$\pm$0.039  & 0.701$\pm$0.052  & - & \bf 0.761$\pm$0.047 \\
HF (4261/684) & 0.877$\pm$0.008  & 0.839$\pm$0.010  & 0.863$\pm$0.018  & - & \bf 0.896$\pm$0.007 \\
Drusen (3990/399) & 0.762 $\pm$0.024  & 0.731$\pm$0.024  & 0.766$\pm$0.029  & - & \bf 0.775$\pm$0.038 \\
RPD (1620/146) & 0.291$\pm$0.044  & 0.302$\pm$0.036  & 0.288$\pm$0.069  & - & \bf 0.335$\pm$0.077 \\
ERM (4338/670) & 0.885$\pm$0.009  & 0.849$\pm$0.014  & 0.850$\pm$0.022  & - & \bf 0.903$\pm$0.010 \\
GA (897/67) & \bf 0.557$\pm$0.063  & 0.234$\pm$0.047  & 0.330$\pm$0.049  & - & 0.556$\pm$0.057 \\
ORA (1999/84) & \bf 0.151$\pm$0.018  & 0.105$\pm$0.008  & 0.143$\pm$0.025  & - & 0.131$\pm$0.019 \\
IRC (3097/553) & 0.932$\pm$0.006  & 0.880$\pm$0.011  & 0.928$\pm$0.012  & -  & \bf 0.940 $\pm$0.006 \\
FPED (3654/387) & 0.931$\pm$0.007  & 0.920 $\pm$0.008  & 0.936$\pm$0.009  & -  & \bf 0.949$\pm$0.006 \\
\midrule
mAP (micro) & 0.814$\pm$0.006  & 0.768$\pm$0.012  & 0.794$\pm$0.010  & 0.599$\pm$0.003 & \bf 0.834$\pm$0.012 \\
mAP (macro) &  0.702$\pm$0.008  & 0.645$\pm$0.009  & 0.680$\pm$0.012  & 0.523$\pm$0.006 & \bf 0.723$\pm$0.014 \\
EMR* & 0.423$\pm$0.015  & 0.164$\pm$0.048  & 0.413$\pm$0.019  & \bf 0.440$\pm$0.003 & 0.438$\pm$0.011 \\
F1* & 0.676$\pm$0.006  & 0.502$\pm$0.024  & 0.676$\pm$0.011  & 0.649$\pm$0.013 & \bf 0.694$\pm$0.009 \\
\bottomrule
\end{tabular}}
\caption{Experimental results comparing our proposed method to other approaches. The per-biomaker scores are shown as mean Average Precision (mAP). The training and test label occurrence is stated beside the biomarker name (training/test). (*) threshold taken at the max F1 score.}
\label{table:table-results}
\end{table}

\section{Conclusion}
\label{sec:conc}
We have presented a novel method to identify pathological biomarkers in OCT slices. Our approach involves detecting biomarkers first slice by slice in the OCT volume and then using a bidirectional LSTM to coherently adjust predictions. As far as we are aware, we are the first to demonstrate that such fine-grained biomarker detection can be achieved in the context of retinal diseases. We have shown that our approach performs well on a substantial patient dataset outperforming other common fusion methods. Future efforts will be focused on extending these results to infer pixel-wise segmentations of found biomarkers relying solely on the per-image labels.

\section*{Acknowledgements}
This work received partial financial support from the Innosuisse Grant \#6362.1 PFLS-LS.
{
\small
\bibliographystyle{splncs}
%\bibliography{string,vision,learning,biomed,optim}
\bibliography{paper1547_top}

\begin{thebibliography}{10}

\bibitem{Bourne2017}
Bourne, R.,  et~al.:
\newblock Magnitude, temporal trends, and projections of the global prevalence
  of blindness and distance and near vision impairment: a systematic review and
  meta-analysis.
\newblock Lancet Global Health \textbf{5} (2017)  e888 -- e897

\bibitem{Apostolopoulos2017}
Apostolopoulos, S., {De Zanet}, S., Ciller, C., Wolf, S., Sznitman, R.:
\newblock {Pathological OCT Retinal Layer Segmentation Using Branch Residual
  U-Shape Networks}.
\newblock In: Medical Image Computing and Computer-Assisted Intervention.
  (2017)  294--301

\bibitem{Hussain2017}
Hussain, M.A., Bhuiyan, A., Turpin, A., Luu, C.D., Smith, R.T., Guymer, R.H.,
  Kotagiri, R.:
\newblock {Automatic Identification of Pathology-Distorted Retinal Layer
  Boundaries Using SD-OCT Imaging}.
\newblock IEEE Transactions on Biomedical Engineering \textbf{64}(7) (2017)
  1638--1649

\bibitem{Roy2017}
Roy, A.G., Conjeti, S., Karri, S.P.K., Sheet, D., Katouzian, A., Wachinger, C.,
  Navab, N.:
\newblock {ReLayNet: retinal layer and fluid segmentation of macular optical
  coherence tomography using fully convolutional networks}.
\newblock Biomedical Optics Express \textbf{8}(8) (2017)

\bibitem{Zhao2017}
Zhao, R., Camino, A., Wang, J., Hagag, A.M., Lu, Y., Bailey, S.T., Flaxel,
  C.J., Hwang, T.S., Huang, D., Li, D., Jia, Y.:
\newblock {Automated drusen detection in dry age-related macular degeneration
  by multiple-depth,en faceoptical coherence tomography.}
\newblock Biomedical optics express \textbf{8}(11) (2017)  5049--5064

\bibitem{Bogunovic2019-nu}
Bogunovic, H.,  et~al.:
\newblock {RETOUCH} -the retinal {OCT} fluid detection and segmentation
  benchmark and challenge.
\newblock IEEE Trans. Med. Imaging (February 2019)

\bibitem{10.1007/11550907_126}
Graves, A., Fern{\'a}ndez, S., Schmidhuber, J.:
\newblock Bidirectional lstm networks for improved phoneme classification and
  recognition.
\newblock In: Artificial Neural Networks: Formal Models and Their Applications.
  (2005)  799--804

\bibitem{Yu2017-gu}
Yu, F., Koltun, V., Funkhouser, T.:
\newblock Dilated residual networks.
\newblock (May 2017)

\bibitem{szummer2008}
Szummer, M., Kohli, P., Hoiem, D.:
\newblock Learning crfs using graph cuts.
\newblock Volume 5303. (10 2008)  582--595

\bibitem{ILSVRC15}
Russakovsky, O., Deng, J., Su, H., Krause, J., Satheesh, S., Ma, S., Huang, Z.,
  Karpathy, A., Khosla, A., Bernstein, M., Berg, A.C., Fei-Fei, L.:
\newblock {ImageNet Large Scale Visual Recognition Challenge}.
\newblock International Journal of Computer Vision \textbf{115}(3) (2015)
  211--252

\end{thebibliography}
}

\end{document}